\documentclass[runningheads]{llncs}
\usepackage[T1]{fontenc}
\usepackage{graphicx}

\usepackage{amssymb}
\usepackage{amsmath}
\usepackage{booktabs}
\usepackage[normalem]{ulem}
\usepackage{multirow}
\usepackage{url}
\usepackage{doi}
\usepackage{hyperref}
\usepackage{subcaption}
\usepackage{seqsplit}
\begin{document}
\title{RelBall: Relation Ball with Quaternion Rotation for Knowledge Graph Completion}
\titlerunning{RelBall} 
%
%


\author{Yike Liu \and
Peijia Xie \and
Chao He \and
Huiling Zhu\thanks{Corresponding author.}}
\authorrunning{Y. Liu et al.}
\institute{South China Normal University, Guangzhou, China}

\maketitle              
\begin{abstract}
Real-world knowledge graphs are often incomplete, lacking many valid facts. Knowledge Graph Completion (KGC) aims to predict missing links using known triples, thereby enhancing graph coverage. A key challenge is modeling diverse relational patterns such as symmetry, antisymmetry, inversion, composition and semantic hierarchy. Existing models such as RotatE can capture symmetric, antisymmetric, inverse, and commutative composition patterns, yet struggle with non-commutative composition. Rotate3D addresses this by introducing non-commutativity via three-dimensional rotations, but still fails to capture the semantic hierarchies prevalent in knowledge graphs. Moreover, both models cannot effectively model one-to-many relations. To overcome these limitations, we propose RelBall, which extends Rotate3D with two innovations. First, our model introduces modulus transformation to model hierarchies, driving abstract concepts toward smaller moduli and concrete instances toward larger ones. Second, it introduces a tail-centric relation ball to model one-to-one, one-to-many, many-to-one, and many-to-many relations. RelBall offers the following advantages: (1) coverage of all relational patterns, including the ones mentioned above; (2) an interpretable hierarchical representation where the modulus directly reflect semantic levels; (3) support for one-to-one, one-to-many, many-to-one, and many-to-many relations. Experiments on multiple datasets demonstrate RelBall's competitive link prediction performance against various baselines.

\end{abstract}
%
%
%

\section{Introduction}\label{}

Knowledge graphs (KGs) represent real-world facts as triples (head entity, relation, tail entity) and are widely used in tasks such as question answering and recommendation systems~\cite{h_ren_association_2023}. However, KGs are often incomplete, with many missing links that degrade downstream performance. Knowledge Graph Completion (KGC) has thus emerged to predict missing entities or relations, improving KG coverage and accuracy.

Knowledge Graph Embedding (KGE) is a prevalent KGC approach that maps entities and relations into a low-dimensional vector space while preserving their structural and semantic information. KGE methods can be categorized into geometric transformations, tensor decompositions, and neural networks. Among these, geometric transformation models have drawn considerable attention due to their parameter efficiency, fast convergence, and high interpretability.

Relations in knowledge graphs exhibit diverse patterns, including symmetry, antisymmetry, inversion, and composition. Composition relations can be further divided into commutative and non-commutative types. For instance, ``wife's father'' and ``father's wife'' refer to father-in-law and mother, respectively---their meanings depend on composition order, underscoring the importance of non-commutativity. While models like TransE and RotatE handle commutative composition well, they struggle with non-commutative cases and often fail to distinguish queries with different composition orders.

Quaternion space, as a hypercomplex space, naturally captures 3D rotational transformations with greater flexibility. Models such as QuatE~\cite{zhang_quaternion_2019}, QuatRE~\cite{nguyen_quatre_2022}, and Rotate3D~\cite{gao_rotate3d_2020} employ quaternions for relational rotation and achieve strong completion performance. However, these methods typically use quaternions solely for modeling relation directions, overlooking the expressive power of the modulus. This limitation makes it difficult to capture the hierarchical structures common in knowledge graphs. Moreover, prior rotation-based models such as RotatE and Rotate3D often assume one-to-one mappings and cannot effectively model one-to-many, many-to-one, or many-to-many relations, which are prevalent in real-world KGs.

To address these limitations, we propose RelBall, a hierarchy-aware knowledge graph embedding model based on quaternion rotation and modulus scaling. RelBall introduces two key innovations. First, it performs rotations in quaternion space to model relational patterns while adjusting hierarchical distances via modulus scaling, enabling simultaneous capture of both semantic hierarchy and relational direction. Second, it introduces a tail-centric relation ball that maps entities onto ball centered at the tail entity, allowing the model to naturally handle one-to-one, one-to-many, many-to-one, and many-to-many relations without additional parameters. The main contributions are as follows:

\begin{enumerate}
\itemsep=0pt
\item \textbf{Semantic Hierarchy Modeling:} Combines quaternion rotation with modulus scaling to model hierarchical relations. Theoretical analysis proves that the modulus directly reflects semantic level, where higher-level entities converge toward smaller moduli and lower-level entities toward larger ones, aligning with human cognitive abstraction.
\item \textbf{Advantages of 3D Rotations:} Leverages quaternions for three-dimensional rotations, avoiding the gimbal lock problem of Euler angles and providing a more stable and expressive rotation representation.
\item \textbf{Relation Ball for Complex Mappings:} Introduces a tail-centric relation ball that enables flexible modeling of one-to-one, one-to-many, many-to-one, and many-to-many relations, overcoming a key limitation of prior rotation-based models.
\item \textbf{Empirical Results:} Experiments on two standard benchmarks demonstrate RelBall's advantages in both hierarchical relation modeling and link prediction against competitive baselines.
\end{enumerate}

This paper is organized as follows. Section \ref{sec:related} reviews related work on knowledge graph embedding and hierarchical relation modeling. Section \ref{sec:preliminaries} covers preliminaries on quaternion rotation and Rotate3D. Section \ref{sec:method} presents the RelBall model along with its theoretical analysis. Section \ref{sec:experiments} reports experimental results and discussion. Section \ref{sec:conclusion} concludes the paper.




\section{Related Work}\label{sec:related}

\subsection{Knowledge Graph Embedding Methods}

Knowledge graph embedding methods can be grouped into three categories based on their modeling approach: geometric transformation models, tensor decomposition models, and neural network models.

\textbf{Geometric transformation models} treat relations as geometric operations in the embedding space. 
Translation-based models like TransE~\cite{bordes_translating_2013} view relations as translations from head to tail entities but struggle with complex relations patterns. 
Rotation-based models like RotatE~\cite{sun_rotate_2019} introduce rotational operations in complex space, capable of modeling symmetric, inverse, and other relations; QuatE~\cite{zhang_quaternion_2019} extends this to quaternion hypercomplex space, and HRotatE~\cite{shah_hrotate_2021} further introduces hierarchy-aware rotations. 
Hyperbolic geometry models like MuRP~\cite{balazevic_multi-relational_2019} and HAKE~\cite{zhang_learning_2020} utilize the exponential growth property of hyperbolic space to model hierarchical structures. BoxE~\cite{abboud_boxe_2020} uses box regions to represent entities and relations, modeling complex relations patterns.

\textbf{Tensor decomposition models} treat the knowledge graph as a third-order tensor for decomposition. DistMult~\cite{yang_embedding_2015} uses diagonal matrices for simplified decomposition but can only handle symmetric relations; ComplEx~\cite{trouillon_complex_2016} introduces complex embeddings, effectively modeling asymmetric relations.

\textbf{Neural network models} leverage deep networks to learn interaction features between entities and relations. ConvE~\cite{dettmers_convolutional_2018} uses 2D convolutions to learn embedding interactions; A2N~\cite{bansal_a2n_2019} introduces attention mechanisms for neighbor aggregation.

\subsection{Hierarchical Relation Modeling}

Modeling hierarchical relations is crucial in knowledge graph embedding. 
Early methods like TransE~\cite{bordes_translating_2013} struggle with this because translation cannot capture semantic distance variations. HAKE~\cite{zhang_learning_2020} uses the magnitude in plane polar coordinates to represent semantic hierarchy.
\subsection{Quaternions in Representation Learning}

Quaternions have also found applications in knowledge graph completion:

QuatE~\cite{zhang_quaternion_2019} first introduced quaternions to knowledge graph embeddings, using the Hamilton product for relational rotations. 
QuatRE~\cite{nguyen_quatre_2022} extended this idea with relation-aware quaternion transformations. 
HRotatE~\cite{shah_hrotate_2021} introduced hierarchy-aware rotations in quaternion space but still focused on rotation operations.

\section{Preliminaries}
\label{sec:preliminaries}

\subsection{Quaternion Basics}

A quaternion $q$ has the following form: $q = a + bi + cj + dk$, where $a, b, c, d \in \mathbb{R}$. 
It consists of one real part and three imaginary parts. 
Complex numbers can be seen as a special case of quaternions with $c = d = 0$. 
A quaternion~\cite{sarabandi_survey_2019} can also be represented as a scalar-vector pair $q = [a, \mathbf{u}]$, where $a \in \mathbb{R}$ and $\mathbf{u} \in \mathbb{R}i + \mathbb{R}j + \mathbb{R}k$.

The imaginary units $i, j, k$ satisfy the following rules:
\begin{equation}
i^2 = j^2 = k^2 = ijk = -1
\end{equation}

From these rules, the following can be obtained:
\begin{equation}
ij = k = -ji, \quad jk = i = -kj, \quad ki = j = -ik
\end{equation}

Based on these equations, the multiplication of two quaternions $q_1 = [a_1, \mathbf{u}_1]$ and $q_2 = [a_2, \mathbf{u}_2]$ is:
\begin{equation}
q_1 q_2 = [a_1 a_2 - \mathbf{u}_1 \cdot \mathbf{u}_2, \ a_1 \mathbf{u}_2 + a_2 \mathbf{u}_1 + \mathbf{u}_1 \times \mathbf{u}_2]
\end{equation}
Importantly, due to the presence of the vector cross product $\mathbf{u}_1 \times \mathbf{u}_2$, multiplication is non-commutative: $q_1 q_2 \neq q_2 q_1$.

\subsection{Representing 3D Rotations with Quaternions}

Let $\mathbf{v} \in \mathbb{R}^3$ denote a point or vector in three-dimensional space~\cite{gao_rotate3d_2020}. To perform rotation operations using quaternions, v is represented as a pure imaginary quaternion with zero real part:
\begin{equation}
v = [0, \mathbf{v}]
\end{equation}

A rotation is defined by an axis and an angle. Let $\mathbf{u} \in \mathbb{R}^3$ be a unit vector representing the direction of the rotation axis, and let $\theta \in \mathbb{R}$ be the rotation angle. The corresponding rotation is represented by the unit quaternion:
\begin{equation}
q = \cos\frac{\theta}{2} + \mathbf{u}\sin\frac{\theta}{2} = \left[\cos\frac{\theta}{2},\; \sin\frac{\theta}{2}\,\mathbf{u}\right]
\end{equation}
This quaternion satisfies $qq^* = 1$, where $q^* = \cos\frac{\theta}{2} - \mathbf{u}\sin\frac{\theta}{2}$ is the conjugate and inverse $q^{-1}$ of $q$.
The rotated vector $\mathbf{v}'$ is obtained via the following quaternion multiplication:
\begin{equation}
v' = q \, v \, q^{-1}
\end{equation}
with $v' = [0, \mathbf{v}']$.

Geometrically, decompose $\mathbf{v}$ into components parallel and perpendicular to the rotation axis $\mathbf{u}$:
\begin{equation}
v = v_{\parallel} + v_{\perp}
\end{equation}
Under the rotation $v' = q v q^{-1}$, the parallel component $\mathbf{v}_{\parallel}$ remains unchanged, while the perpendicular component $\mathbf{v}_{\perp}$ rotates around $\mathbf{u}$ by angle $\theta$. Therefore, the quaternion multiplication $v' = q v q^{-1}$ exactly represents a rotation of $\mathbf{v}$ by $\theta$ about the axis $\mathbf{u}$.

\subsection{Rotate3D Model}

Rotate3D~\cite{gao_rotate3d_2020} projects entities to $\mathbf{h}, \mathbf{t} \in \mathbb{R}^{3 \times n}$ and defines each relation as an element-wise rotation. For a triple $(h, r, t)$:
\begin{equation}
t^{(i)} = q_i h^{(i)} q_i^{-1}
\end{equation}
where $i \in \{1, \ldots, n\}$, $h^{(i)} = [0, \mathbf{h}^{(i)}]$, $t^{(i)} = [0, \mathbf{t}^{(i)}]$, $q_i = [\cos\frac{\theta}{2},\; \sin\frac{\theta}{2}\,\mathbf{u}]$, and $\theta$ and $\mathbf{u}$ constitute $\mathbf{r}^{(i)}$. For simplicity, this rotation is written as:
\begin{equation}
\mathbf{t}^{(i)} = \mathbf{h}^{(i)} \odot \mathbf{r}^{(i)}
\end{equation}
where $\odot: \mathbb{R}^3 \times \mathbb{R}^3 \to \mathbb{R}^3$ represents a 3D rotation induced by $\mathbf{r}^{(i)}$. This follows the standard Rotate3D formulation. The score function is defined as:
\begin{equation}
f_r(h, t) = -\sum_{i=1}^{n} \| \mathbf{h}^{(i)} \odot \mathbf{r}^{(i)} - \mathbf{t}^{(i)} \|_p
\end{equation}
where $\|\cdot\|_p$ denotes the L1 norm if $p=1$ or the L2 norm if $p=2$. Let $\mathbf{b} \in \mathbb{R}^n$ be a relation-specific bias vector, where $b^{(i)}$ denotes its $i$-th component (a scalar). Then:
\begin{equation}
f_r(h, t) = -\sum_{i=1}^{n} \| (\mathbf{h}^{(i)} \odot \mathbf{r}^{(i)}) \cdot b^{(i)} - \mathbf{t}^{(i)} \|_p
\end{equation}
Rotate3D primarily focuses on rotational transformations. Its relation-specific bias term performs element-wise multiplication on the rotated vector. The original Rotate3D paper does not constrain the sign of this bias term. This paper further imposes a range constraint on the bias. Geometric modeling and experimental results demonstrate that the constrained bias term captures semantic distances in hierarchical relations. This capability is not explored in the original Rotate3D paper.
\section{RelBall Model}
\label{sec:method}

\subsection{Model Formulation}

RelBall extends Rotate3D by introducing modulus scaling and a relation ball mechanism.  
Let $\mathbf{s}_r, \boldsymbol{\rho}_r \in (\mathbb{R}^+)^n$ be learnable vectors with components $s_r^{(i)}$ and $\rho_r^{(i)}$, respectively, where $\mathbf{s}_r$ is the scaling factor and $\boldsymbol{\rho}_r$ is the radius factor.  
The relation ball radius in the $i$-th subspace is proportional to the local modulus of the tail entity:
\begin{equation}
R_r^{(i)} = \rho_r^{(i)} \|\mathbf{t}^{(i)}\|_p,
\end{equation}
where $\|\cdot\|_p$ is the $L_p$-norm used in Rotate3D.

The head entity is first scaled then rotated:
\begin{equation}
\mathbf{t}^{(i)} = (s_r^{(i)} \cdot \mathbf{h}^{(i)}) \odot \mathbf{r}^{(i)},
\end{equation}
with $\odot$ denoting the quaternion rotation defined in Rotate3D. The scoring function becomes:
\begin{equation}
f_r(h, t) = -\sum_{i=1}^{n} \max\bigl(0, \|(s_r^{(i)} \cdot \mathbf{h}^{(i)}) \odot \mathbf{r}^{(i)} - \mathbf{t}^{(i)}\|_p - R_r^{(i)}\bigr).
\end{equation}

\subsection{Geometric Representation for Semantic Hierarchies and Complex Relations}


\begin{figure}
\centering
\includegraphics[width=2.2in]{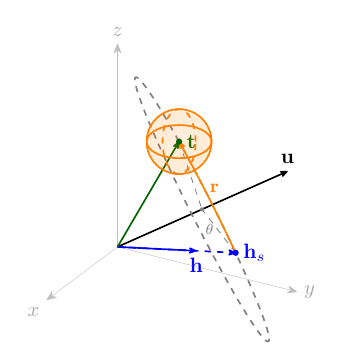}
\caption{RelBall Model Illustration}
\label{fig:rotate3dm2}
\end{figure}

The model leverages radial layering to characterize semantic hierarchies. Entities with smaller moduli correspond to more abstract concepts, whereas those with larger moduli represent concrete ones. Complex relations (1-to-1, 1-to-N, N-to-1, N-to-N) are modeled using relation ball. Specifically, the transformed head entity is constrained within the ball centered at the tail entity, enabling multiple heads to map to one tail, or a single head to belong to multiple relation ball. The overall architecture of RelBall is illustrated in Figure~\ref{fig:rotate3dm2}.

Combining scaling, rotation and relation ball operations, RelBall jointly learns semantic hierarchies and complex relations. Concretely, the head entity $h$ is scaled by factor $s_r$, and then matched with the tail entity $t$ through quaternion rotation.

RelBall integrates quaternion rotation, modulus scaling and relation ball into a unified framework. Rotation captures relational directionality, scaling encodes hierarchical semantics, and relation ball handle complex relations. This framework offers a full geometric explanation for knowledge graph representation learning.

\subsection{Semantic Hierarchy Characteristics}

In the RelBall model, hierarchical structures are primarily represented by the modulus scaling factor $s_r$, while the relation ball radius $R_r$ provides a geometric interpretation of the tolerance boundary for hierarchical mappings. Specifically:

When $s_r < 1$, the head entity $h$ resides at a higher semantic level and the tail entity $t$ at a lower level. If $\rho_r$ is small, the ball radius $R_r$ is correspondingly small, indicating a precise mapping from higher-level to lower-level entities with limited tolerance. If $\rho_r$ is large, greater deviation is permitted, reflecting a degree of fuzziness in the hierarchical mapping.

When $s_r = 1$, the head and tail entities lie at the same level of abstraction. In this case, the magnitude of $\rho_r$ determines the tolerance range between same-level entities: a small $\rho_r$ enforces strict equivalence relations, while a large $\rho_r$ allows for looser semantic similarity.

When $s_r > 1$, the head entity $h$ resides at a lower semantic level and the tail entity $t$ at a higher level. Similarly, $\rho_r$ controls the ball tolerance for mappings from lower-level to higher-level entities: a small $\rho_r$ indicates precise hypernym mappings, while a large $\rho_r$ permits fuzzy hypernym relations.

Furthermore, $s_r$ and $R_r$ admit a joint interpretation: $s_r$ determines the direction and magnitude of the hierarchical shift, while $R_r$ determines the tolerance range of that shift. Together, they enable RelBall to model both strict hierarchical structures (small $s_r$ deviation with small $\rho_r$) and loose hierarchical structures (large $s_r$ deviation with large $\rho_r$).

\subsection{Modeling Capability of Relation Patterns}

RelBall maintains strong modeling capability for relation patterns, representing the following types of relations:

\subsubsection{Symmetric Relations}

For a symmetric relation pattern, if $(h, r, t) \in \mathcal{T}$ and $(t, r, h) \in \mathcal{T}$, then the ball constraints impose:
\[
\begin{cases}
\|t - s_r \cdot (q_r \, h \, q_r^{-1})\|_p \leq R_r \\[6pt]
\|h - s_r \cdot (q_r \, t \, q_r^{-1})\|_p \leq R_r
\end{cases}
\]
Substituting the first inequality into the second yields:
\[
\begin{aligned}
\|h - s_r \cdot (q_r \, t \, q_r^{-1})\|_p
&= \bigl\|h - s_r \cdot \bigl( q_r \, \bigl( s_r \cdot (q_r \, h \, q_r^{-1}) + \delta \bigr) \, q_r^{-1} \bigr) \bigr\|_p \\
&= \bigl\|h - s_r^2 \cdot \bigl( q_r^2 \, h \, q_r^{-2} \bigr) - s_r \cdot (q_r \, \delta \, q_r^{-1}) \bigr\|_p \\
&\leq R_r
\end{aligned}
\]
where $\|\delta\|_p \leq R_r$ is the deviation vector (pure quaternion) from the first ball constraint. For this inequality to hold for arbitrary $h$ within the ball tolerance, the core rotation-scaling condition must be satisfied:
\[
s_r^2 = 1 \quad \text{and} \quad q_r^2 = \pm 1
\]
Since $s_r \in \mathbb{R}^+$, it follows that $s_r = 1$. The condition $q_r^2 = \pm 1$ corresponds to rotation angles $\theta_r \in \{0, \pi\}$.

\subsubsection{Antisymmetric Relations}

For an antisymmetric relation pattern, $(h, r, t) \in \mathcal{T} \Rightarrow (t, r, h) \notin \mathcal{T}$, the ball constraints give:
\[
\begin{cases}
\|t - s_r \cdot (q_r \, h \, q_r^{-1})\|_p \leq R_r \\[6pt]
\|h - s_r \cdot (q_r \, t \, q_r^{-1})\|_p > R_r
\end{cases}
\]
Expanding the second inequality with the first constraint:
\[
\bigl\|h - s_r^2 \cdot \bigl( q_r^2 \, h \, q_r^{-2} \bigr) - s_r \cdot (q_r \, \delta \, q_r^{-1}) \bigr\|_p > R_r
\]
For antisymmetry to be enforced, at least one of the following conditions must hold:
\[
s_r^2 \neq 1 \quad \text{or} \quad q_r^2 \neq \pm 1
\]

\subsubsection{Composition Relations}

For a composition relation pattern, if $(h, r_1, t) \in \mathcal{T}$, $(t, r_2, z) \in \mathcal{T}$ and $(h, r_3, z) \in \mathcal{T}$, then the ball constraints impose:
\[
\begin{cases}
\|t - s_{r_1} \cdot (q_{r_1} \, h \, q_{r_1}^{-1})\|_p \leq R_{r_1} \\[6pt]
\|z - s_{r_2} \cdot (q_{r_2} \, t \, q_{r_2}^{-1})\|_p \leq R_{r_2} \\[6pt]
\|z - s_{r_3} \cdot (q_{r_3} \, h \, q_{r_3}^{-1})\|_p \leq R_{r_3}
\end{cases}
\]
Combining the first two constraints:
\[
\begin{aligned}
z &= s_{r_2} \cdot (q_{r_2} \, t \, q_{r_2}^{-1}) + \delta_2 \\
&= s_{r_2} \cdot \bigl( q_{r_2} \, \bigl( s_{r_1} \cdot (q_{r_1} \, h \, q_{r_1}^{-1}) + \delta_1 \bigr) \, q_{r_2}^{-1} \bigr) + \delta_2 \\
&= (s_{r_1} s_{r_2}) \cdot \bigl( (q_{r_2} q_{r_1}) \, h \, (q_{r_2} q_{r_1})^{-1} \bigr) + s_{r_2} \cdot (q_{r_2} \, \delta_1 \, q_{r_2}^{-1}) + \delta_2
\end{aligned}
\]
where $\|\delta_1\|_p \leq R_{r_1}$ and $\|\delta_2\|_p \leq R_{r_2}$ (both pure quaternions). For $z$ to satisfy the third ball constraint, the ideal composition conditions are:
\[
s_{r_3} = s_{r_1} s_{r_2} \quad \text{and} \quad q_{r_3} = \pm \, q_{r_2} q_{r_1}
\]
Furthermore, the ball radii must satisfy a cumulative constraint to accommodate the propagated errors:
\[
R_{r_3} \geq s_{r_2} \cdot R_{r_1} + R_{r_2}.
\]
If $r_1$ and $r_2$ are commutative, then:
\[
q_{r_1} q_{r_2} = q_{r_2} q_{r_1}
\]
If $r_1$ and $r_2$ are non-commutative, then:
\[
q_{r_1} q_{r_2} \neq q_{r_2} q_{r_1}
\]

\subsubsection{Inverse Relations}

For an inverse relation pattern, if $(h, r_1, t) \in \mathcal{T}$ and $(t, r_2, h) \in \mathcal{T}$, where $r_1$ and $r_2$ are a pair of inverse relations, the ball constraints impose:
\[
\begin{cases}
\|t - s_{r_1} \cdot (q_{r_1} \, h \, q_{r_1}^{-1})\|_p \leq R_{r_1} \\[6pt]
\|h - s_{r_2} \cdot (q_{r_2} \, t \, q_{r_2}^{-1})\|_p \leq R_{r_2}
\end{cases}
\]
Substituting the first inequality into the second yields:
\[
\begin{aligned}
\|h - s_{r_2} \cdot (q_{r_2} \, t \, q_{r_2}^{-1})\|_p
&= \bigl\|h - s_{r_2} \cdot \bigl( q_{r_2} \, \bigl( s_{r_1} \cdot (q_{r_1} \, h \, q_{r_1}^{-1}) + \delta_1 \bigr) \, q_{r_2}^{-1} \bigr) \bigr\|_p \\
&= \bigl\|h - (s_{r_1} s_{r_2}) \cdot \bigl( (q_{r_2} q_{r_1}) \, h \, (q_{r_2} q_{r_1})^{-1} \bigr) - s_{r_2} \cdot (q_{r_2} \, \delta_1 \, q_{r_2}^{-1}) \bigr\|_p \\
&\leq R_{r_2}
\end{aligned}
\]
where $\|\delta_1\|_p \leq R_{r_1}$ is a pure quaternion deviation. For this to hold for arbitrary $h$ within the ball tolerance, the core conditions are:
\[
s_{r_1} s_{r_2} = 1 \quad \text{and} \quad q_{r_2} q_{r_1} = \pm 1
\]
Since $s_{r_1}, s_{r_2} \in \mathbb{R}^+$, it follows that $s_{r_2} = s_{r_1}^{-1}$. The quaternion condition $q_{r_2} q_{r_1} = \pm 1$ implies $q_{r_2} = \pm q_{r_1}^{-1}$, i.e., the rotation of the inverse relation is the inverse (or negative inverse) of the original rotation. Additionally, the ball radii must satisfy:
\[
R_{r_2} \geq s_{r_2} \cdot R_{r_1} = s_{r_1}^{-1} \cdot R_{r_1}.
\]

\subsection{Loss Function and Training}

During training, each triple $(h, r, t)$ in the training set is treated as a positive sample. 
For each positive samples, $n$ negative triples $(h'_i, r, t'_i)$ are constructed by randomly replacing the head or tail entity, ensuring that these negative samples are not present in the training set. 
The model is optimized using the negative sampling loss with self-adversarial training~\cite{sun_rotate_2019}:

\begin{equation}
\begin{split}
L = &-\log \sigma(\gamma - f_r(h, t)) \\
&-\sum_{i=1}^{n} p(h'_i, r, t'_i) \log \sigma(f_r(h'_i, t'_i) - \gamma)
\end{split}
\label{eq:loss_function}
\end{equation}

where $\sigma$ is the sigmoid function, $\gamma$ is a margin hyperparameter, and $f_r(h, t)$ is the score function for the triple. The negative sample sampling probability is defined as:

\begin{equation}
p(h'_j, r, t'_j) = \frac{\exp(\alpha f_r(h'_j, t'_j))}{\sum_{i=1}^{n} \exp(\alpha f_r(h'_i, t'_i))}
\end{equation}

where $\alpha$ is the sampling temperature parameter controlling the sampling weight of negative samples. 
This mechanism assigns higher sampling probabilities to hard negative samples with higher scores, thus improving training efficiency.

\subsection{Implementation Details}

RelBall is implemented using PyTorch with the following main configurations: Hyperparameter ranges include embedding dimension $d \in \{500, 750, 1000, 1500\}$, margin parameter $\gamma \in \{6, 9, 12, 24\}$, self-adversarial sampling temperature $\tau \in \{0.5, 1.0\}$, batch size $B \in \{512, 1024\}$, regularization coefficient $\lambda \in [0, 1]$, $p$-norm $p \in \{1, 2\}$, number of negative samples $N_{\text{neg}} = 256$, maximum training steps $T \in [8\times10^4, 2.0\times10^5]$. The Adam optimizer is used with learning rate $\eta \in [5\times10^{-5}, 2\times10^{-4}]$, and a learning rate decay strategy is employed.

\section{Experiments}
\label{sec:experiments}

\subsection{Experimental Setup}

\subsubsection{Datasets}
Experiments on RelBall are performed on two standard benchmark knowledge graph completion datasets: WN18RR~\cite{toutanova_observed_2015} and FB15k-237~\cite{dettmers_convolutional_2018}.
Table \ref{tab:dataset_stats} summarizes their statistics.

\begin{table}
\caption{Dataset Statistics}
\label{tab:dataset_stats}
\centering
\begin{tabular}{lccccc}
\toprule
\textbf{Dataset} & \textbf{\#Entities} & \textbf{\#Relations} & \textbf{\#Training} & \textbf{\#Validation} & \textbf{\#Test} \\
\midrule
WN18RR & 40943 & 11 & 86835 & 3034 & 3134 \\
FB15k-237 & 14541 & 237 & 272115 & 17535 & 20466 \\
\bottomrule
\end{tabular}
\end{table}

\subsubsection{Baseline Methods}
RelBall is compared against several geometric transformation baselines, including RotatE, QuatE, QuatRE, Rotate3D, HA-RotatE, HAKE, MuRP, HBE, and RotatH. 
Among these, HA-RotatE, HAKE, MuRP, and HBE are hierarchy-aware, 
while RotatH is capable of handling complex relations.

\subsubsection{Evaluation Metrics}
For each triple $(h, r, t)$ in the test set, the head or tail entity is replaced with every candidate entity to create candidate triples. 
The filtered setting is used, excluding existing valid triples. Evaluation metrics include Mean Reciprocal Rank (MRR) and Hits@k (k = 1, 3, 10).

\subsubsection{Implementation Details}
All experiments are run on NVIDIA RTX 4090 GPUs.

\subsection{Main Results}

This section presents the experimental results in four parts. 
First, the overall performance of RelBall is compared with baseline models. 
Second, the distribution of scaling factor values across different relations is analyzed. 
Third, the performance of RelBall on complex relations is evaluated. 
Finally, the contribution of each component is assessed through ablation studies.

\subsubsection{Overall Performance}

\begin{table*}[b!]
\centering
\caption{Link prediction results on WN18RR and FB15k-237.}
\label{tab:main_results2}
\resizebox{\textwidth}{!}{%
\begin{tabular}{lcccccccc}
\toprule
\multirow{2}{*}{Model} & \multicolumn{4}{c}{WN18RR} & \multicolumn{4}{c}{FB15k-237} \\
\cmidrule(lr){2-5}\cmidrule(lr){6-9}
 & MRR & H@1 & H@3 & H@10 & MRR & H@1 & H@3 & H@10 \\
\midrule
HAKE~\cite{zhang_learning_2020} & \textbf{0.497} & \textbf{0.452} & \textbf{0.516} & \textit{0.582} & \textit{0.346} & \uline{0.250} & 0.381 & \textit{0.542} \\
HA-RotatE~\cite{wang_hierarchical-aware_2021} & \uline{0.491} & 0.445 & \uline{0.511} & \textbf{0.587} & \textit{0.346} & \uline{0.250} & \textit{0.384} & 0.539 \\
MuRP~\cite{balazevic_multi-relational_2019} & 0.481 & 0.440 & 0.495 & 0.566 & 0.335 & \textit{0.243} & 0.367 & 0.518 \\
HBE~\cite{pan_hyperbolic_2021} & 0.488 & \uline{0.448} & 0.502 & 0.570 & 0.336 & 0.239 & 0.372 & 0.534 \\
RotatH~\cite{trouillon_complex_2016} & 0.472 & 0.432 & 0.484 & 0.554 & 0.342 & \uline{0.250} & 0.378 & 0.526 \\
RotatE~\cite{sun_rotate_2019} & 0.476 & 0.428 & 0.492 & 0.571 & 0.338 & 0.241 & 0.375 & 0.533 \\
QuatE~\cite{zhang_quaternion_2019} & 0.481 & 0.436 & 0.500 & 0.564 & 0.311 & 0.221 & 0.342 & 0.495 \\
QuatRE~\cite{nguyen_quatre_2022} & 0.479 & 0.429 & 0.503 & 0.571 & 0.332 & 0.238 & 0.367 & 0.522 \\
Rotate3D~\cite{gao_rotate3d_2020} & \textit{0.489} & 0.442 & 0.505 & 0.579 & \uline{0.347} & \uline{0.250} & \uline{0.385} & \uline{0.543} \\
\midrule
RelBall & \uline{0.491} & \textit{0.446} & \textit{0.507} & \uline{0.583} & \textbf{0.349} & \textbf{0.251} & \textbf{0.387} & \textbf{0.544} \\
\bottomrule
\end{tabular}%
}
\end{table*}

Based on the experimental results presented in Table~\ref{tab:main_results2}, the following conclusions can be drawn. This table compares models without explicit semantic hierarchy modeling capabilities (RotatE, QuatE, QuatRE, Rotate3D), aiming to verify the fundamental effectiveness of RelBall in general knowledge graph reasoning tasks. It also compares models with semantic hierarchy awareness or complex relations modeling capabilities (HAKE, HA-RotatE, MuRP, HBE, RotatH), to evaluate the competitive advantage of RelBall in handling complex semantic hierarchical relations.

On the FB15k-237 dataset, RelBall demonstrates excellent modeling capability, ranking first across all evaluation metrics, which fully demonstrates its comprehensive advantage in handling complex relations patterns contained in this dataset. On the WN18RR datasets, RelBall also exhibits outstanding robustness, with all evaluation metrics ranking among the top. These results confirm its strong generalization ability and robustness.

\subsubsection{Semantic Hierarchy Analysis}

\begin{figure}[b!]
    \centering
    \begin{subfigure}{0.48\textwidth}
        \centering
        \includegraphics[width=\textwidth]{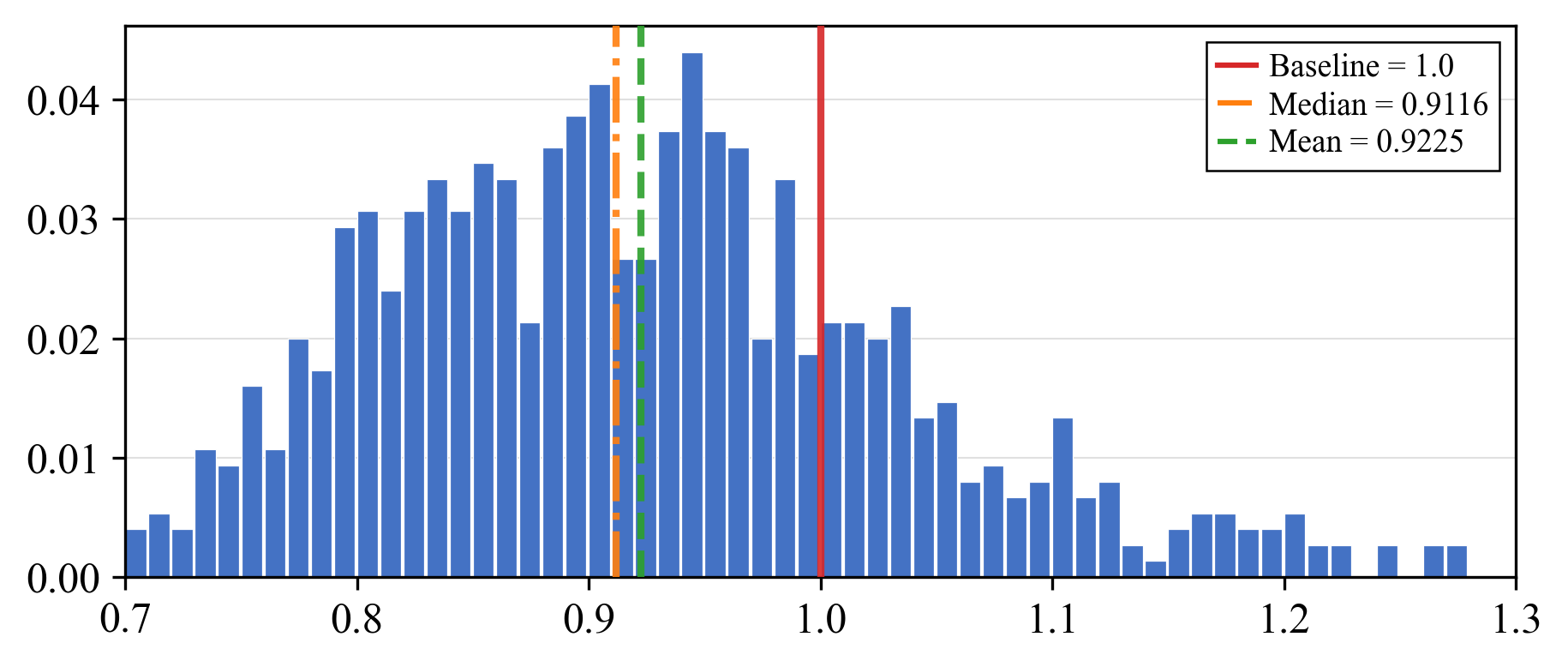}
        \caption{\_hypernym}
        \label{fig:dist_hypernym}
    \end{subfigure}
    \hfill
    \begin{subfigure}{0.48\textwidth}
        \centering
        \includegraphics[width=\textwidth]{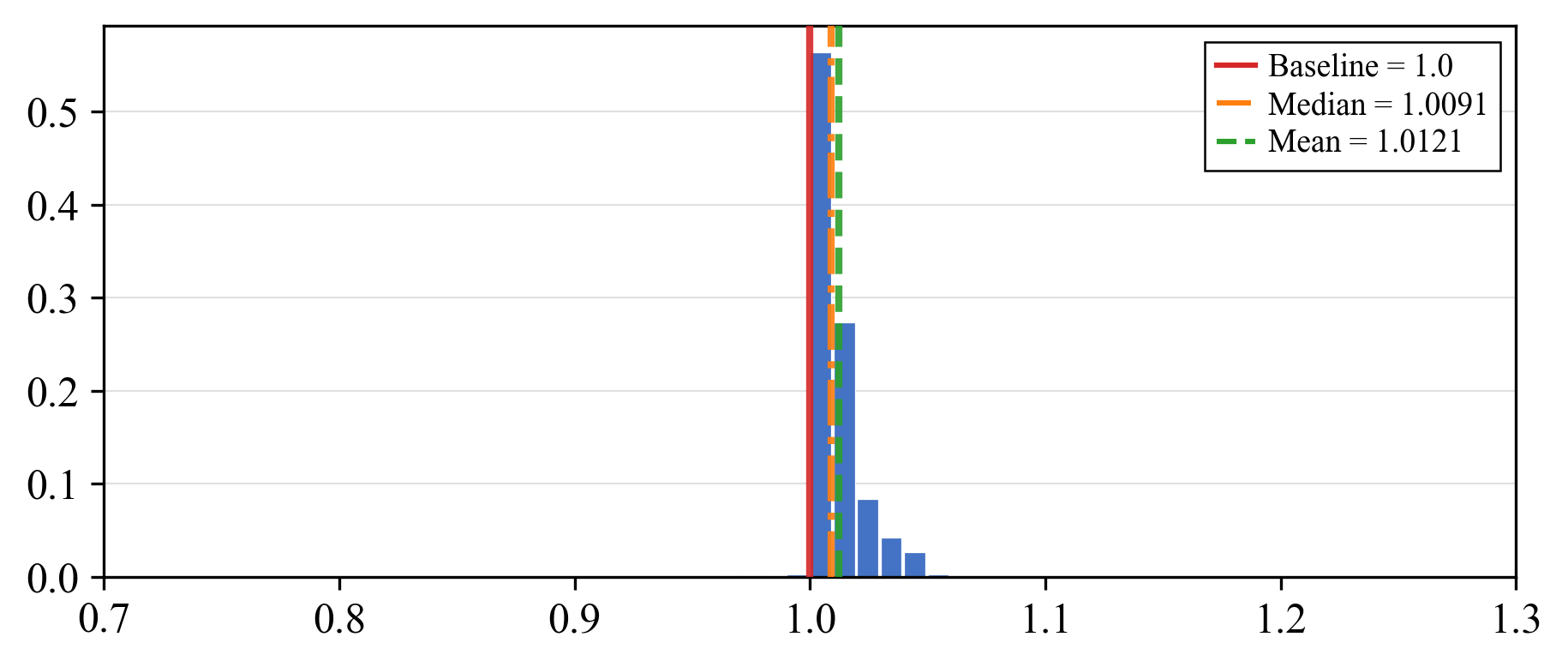}
        \caption{\_verb\_group}
        \label{fig:dist_verbgroup}
    \end{subfigure}
    
    \vspace{4pt}
    
    \begin{subfigure}{0.48\textwidth}
        \centering
        \includegraphics[width=\textwidth]{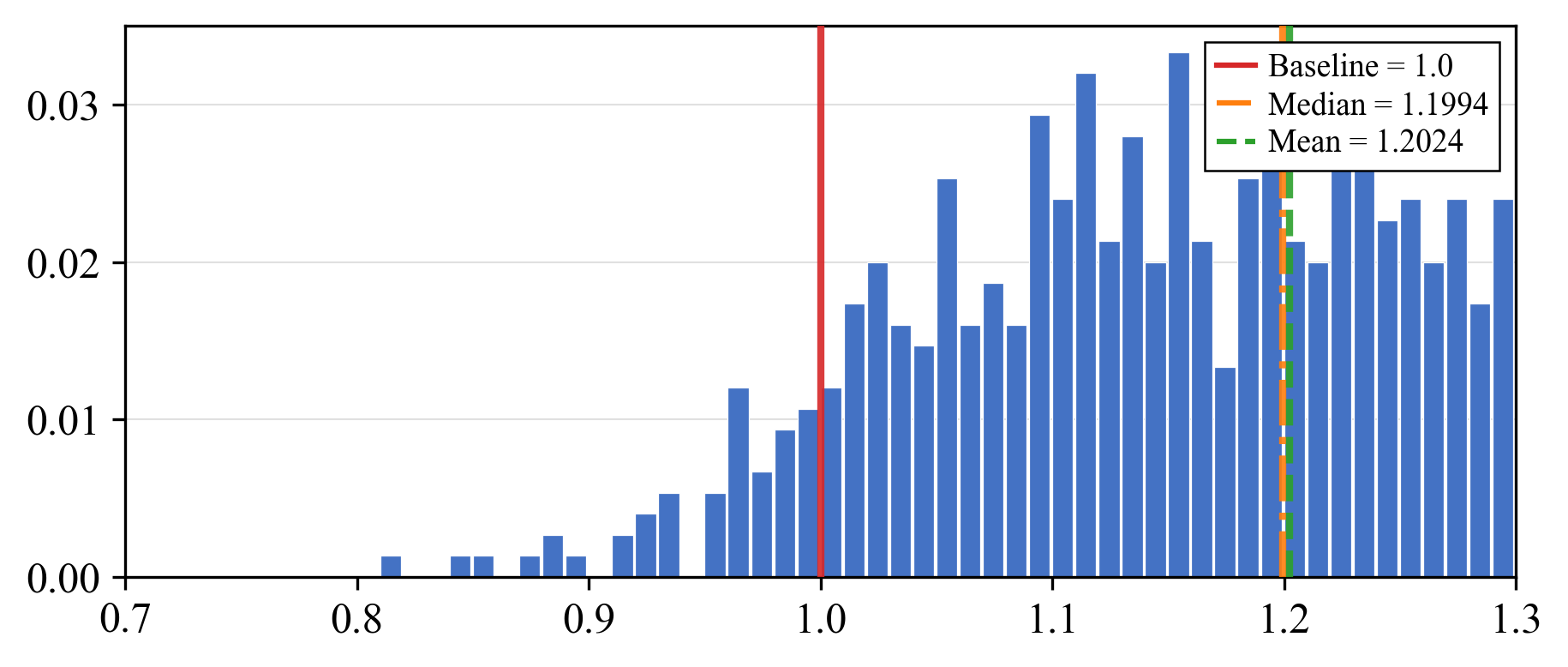}
        \caption{\_member\_meronym}
        \label{fig:dist_meronym}
    \end{subfigure}
    
    \caption{Distribution of the modulus scale parameter across three relation types. 
             Red solid line: Baseline ($=1.0$); green dashed line: Mean; 
             orange dash-dotted line: Median.}
    \label{fig:scale_distribution}
\end{figure}

This paper analyzes the distribution of scaling factor values across different relations.

In Figure \ref{fig:scale_distribution}, both the mean and median scaling factors for subfigures (a), fall below 1; for (b), they approximate 1; for (c), they exceed 1. 
This aligns with theoretical expectations: as discussed in the hierarchical interpretation of the modulus transformation, the scaling factor mathematically captures semantic hierarchical structures, where values below 1, near 1, and above 1 respectively correspond to the head entity being at a higher, same, or lower semantic level than the tail entity.

\subsubsection{Performance by Relation Type}

\begin{table*}[htbp]
\centering
\small
\caption{The results of mapping properties of relations on FB15k-237. [$\dagger$] denotes results reported in~\cite{zhang_knowledge_2022}. [$\ddagger$] denotes results from~\cite{trouillon_complex_2016}.}
\label{tab:relation_breakdown}
\setlength{\tabcolsep}{4pt}
\begin{tabular}{lcccccccc}
\toprule

\multirow{2}{*}{Category} & \multicolumn{4}{c}{Predicting Head (MRR)} & \multicolumn{4}{c}{Predicting Tail (MRR)} \\
\cmidrule(lr){2-5} \cmidrule(lr){6-9}
 & 1-to-1 & 1-to-N & N-to-1 & N-to-N & 1-to-1 & 1-to-N & N-to-1 & N-to-N \\
\midrule
TransE~\cite{bordes_translating_2013}[$\dagger$] & \textit{0.490} & 0.450 & 0.081 & 0.248 & 0.481 & 0.070 & 0.746 & 0.364 \\
ComplEx~\cite{trouillon_complex_2016}[$\dagger$] & 0.367 & \textit{0.463} & \textit{0.091} & 0.249 & 0.362 & 0.064 & 0.737 & 0.357 \\
RotatE~\cite{sun_rotate_2019}[$\dagger$] & \uline{0.496} & \textit{0.463} & 0.086 & 0.253 & \textit{0.482} & \textit{0.072} & \uline{0.757} & \textit{0.369} \\
ReflectE~\cite{zhang_knowledge_2022}[$\dagger$] & 0.459 & 0.448 & \textbf{0.137} & \textbf{0.282} & 0.466 & \textbf{0.085} & \textit{0.753} & \textbf{0.392} \\
RotatH~\cite{trouillon_complex_2016}[$\ddagger$] & 0.481 & \uline{0.464} & 0.077 & \textit{0.268} & \uline{0.484} & 0.071 & 0.744 & \uline{0.381} \\
\midrule
RelBall & \textbf{0.504} & \textbf{0.480} & \uline{0.117} & \uline{0.269} & \textbf{0.486} & \uline{0.075} & \textbf{0.777} & \uline{0.381} \\
\bottomrule

\multirow{2}{*}{Category} & \multicolumn{4}{c}{Predicting Head (Hits@10)} & \multicolumn{4}{c}{Predicting Tail (Hits@10)} \\
\cmidrule(lr){2-5} \cmidrule(lr){6-9}
 & 1-to-1 & 1-to-N & N-to-1 & N-to-N & 1-to-1 & 1-to-N & N-to-1 & N-to-N \\
\midrule
TransE~\cite{bordes_translating_2013}[$\dagger$] & \uline{0.594} & \uline{0.659} & 0.162 & 0.458 & \uline{0.583} & \textit{0.138} & 0.879 & 0.606 \\
ComplEx~\cite{trouillon_complex_2016}[$\dagger$] & 0.521 & 0.655 & \textit{0.170} & 0.454 & 0.531 & 0.126 & 0.862 & 0.591 \\
RotatE~\cite{sun_rotate_2019}[$\dagger$] & \uline{0.594} & \textit{0.658} & 0.167 & 0.463 & 0.563 & 0.131 & \textit{0.880} & \textit{0.609} \\
ReflectE~\cite{zhang_knowledge_2022}[$\dagger$] & \textit{0.563} & 0.642 & \textbf{0.248} & \textbf{0.484} & \textit{0.568} & \textbf{0.157} & \uline{0.883} & \textbf{0.623} \\
RotatH~\cite{trouillon_complex_2016}[$\ddagger$] & 0.542 & 0.656 & 0.148 & \textit{0.476} & \textit{0.568} & 0.128 & 0.874 & 0.607 \\
\midrule
RelBall & \textbf{0.609} & \textbf{0.679} & \uline{0.220} & \uline{0.483} & \textbf{0.599} & \uline{0.145} & \textbf{0.890} & \uline{0.619} \\
\bottomrule

\end{tabular}
\end{table*}

Furthermore, RelBall is evaluated on different relation types, including 1-to-1, 1-to-N, N-to-1, and N-to-N. Specifically, head prediction and tail prediction tasks are analyzed separately. The head prediction is processed by taking each positive triple, removing the head entity, and keeping the relation and tail entity. The head is then replaced with candidate entities to predict whether the newly created triple is valid, i.e., predicting $(?, r, t)$. Tail prediction is performed in the same way, but the replaced entity is the tail entity, i.e., predicting $(h, r, ?)$. The results are summarized in Table~\ref{tab:relation_breakdown}. 
As shown in the table, RelBall achieves particularly strong performance on complex relations. 
These gains can be attributed to the ball constraint, which allows multiple head or tail entities to be valid within the tolerance boundary defined by $R_r$.

\subsubsection{Ablation Study}
\begin{table}[htbp]
\centering
\caption{Ablation study on model components on WN18RR and FB15k-237.}
\label{tab:ablation}
\setlength{\tabcolsep}{3pt}
\begin{tabular}{cccccccccc}
\toprule
\multirow{2}{*}{$s_r$} & \multirow{2}{*}{$\rho_r$} & \multicolumn{4}{c}{WN18RR} & \multicolumn{4}{c}{FB15k-237} \\
\cmidrule(lr){3-6} \cmidrule(lr){7-10}
 & & MRR & Hit@1 & Hit@3 & Hit@10 & MRR & Hit@1 & Hit@3 & Hit@10 \\
\midrule
\checkmark & \checkmark & \uline{0.491} & \textbf{0.446} & \uline{0.507} & \textbf{0.583} & \textbf{0.349} & \textbf{0.251} & \textbf{0.387} & \textbf{0.544} \\
\checkmark   & $\times$   & \textbf{0.493} & \textbf{0.446} & \textbf{0.510} & \textbf{0.583} & \uline{0.347} & \uline{0.250} & \uline{0.385} & \uline{0.543} \\

$\times$   & \checkmark & 0.486 & \uline{0.441} & 0.503 & \textit{0.574} & \textit{0.344} & \textit{0.248} & \textit{0.382} & \textit{0.537} \\
$\times$ & $\times$   & \textit{0.487} & \uline{0.441} & \textit{0.504} & \uline{0.577} & \textit{0.344} & \textit{0.248} & \textit{0.382} & \textit{0.537} \\

\bottomrule
\end{tabular}
\vspace{4pt}
\begin{minipage}{\columnwidth}
\footnotesize
Note: $\checkmark$ indicates the component is included, $\times$ indicates it is removed. The four rows represent: (1) with both scale and radius; (2) with scale only; (3) with radius only; (4) without both.
\end{minipage}
\end{table}

To evaluate the contribution of each component in RelBall, ablation experiments are conducted on WN18RR and FB15k-237 by selectively removing the modulus scaling factor $s_r$ and the radius factor $\rho_r$. 
The modulus scaling factor $s_r$ controls the semantic-level adjustment of the head entity, while the radius factor $\rho_r$ determines the size of the relational ball centered at the tail entity. 
Removing $s_r$ reduces the modulus transformation to an identity mapping, restricting the model to represent relations only at the same semantic level. 
Removing $\rho_r$ collapses the relational ball into a single point, eliminating tolerance for complex relations. 

The results are reported in Table~\ref{tab:ablation}. The results show that the complete RelBall model, which includes both $s_r$ and $\rho_r$, achieves excellent performance on most metrics.

\section{Conclusion}
\label{sec:conclusion}

This paper proposes RelBall, a knowledge graph embedding model that integrates quaternion rotation, modulus transformation, and relation ball. 
Theoretical analysis and experimental results demonstrate that modulus transformation reflects the semantic hierarchy of entities: higher-level entities correspond to smaller moduli, while lower-level entities correspond to larger moduli. 
Meanwhile, the introduction of the relation ball enables it to effectively model complex relations.

%
%
%
\bibliographystyle{splncs04}
\bibliography{mybibliography}

\end{document}